%% file: document.tex
\documentclass{article}

\usepackage{arxiv}
\usepackage{ dsfont }
\usepackage[utf8]{inputenc} 
\usepackage[T1]{fontenc}    
\usepackage{hyperref}       
\usepackage{url}            
\usepackage{booktabs}       
\usepackage{amsfonts}       
\usepackage{nicefrac}       
\usepackage{microtype}      
\usepackage{lipsum}
\usepackage{graphicx}
\usepackage{booktabs}
\usepackage{longtable}

\title{Effects of sparse rewards of different magnitudes in the speed of learning of model-based actor critic methods}

\input{Sections/author.tex}

\begin{document}
\maketitle

\input{Sections/abstract.tex}

\input{Sections/keywords.tex}

\input{Sections/introduction.tex}

\input{Sections/methods.tex}

\input{Sections/results.tex}

\input{Sections/conclusion.tex}

\nocite{bib:memory_replay_effects}
\nocite{bib:Experience_Replay}
\nocite{bib:dpg_silver2014}
\nocite{bib:off_policy_ac_degris2012}
\nocite{bib:apprentice_learning_irl}
\nocite{bib:rafati2019learning}
\nocite{bib:lanier2019curiositydriven}
\bibliographystyle{unsrt}  
\bibliography{Sections/references}  

\input{Sections/supplemental.tex}

\end{document}

%% file: Sections/author.tex
\author{
  Juan ~Vargas \\
  Department of Mechanical Engineering\\
  Carnegie Mellon University\\
  Pittsburgh, PA 15213 \\
  \texttt{jvargass@andrew.cmu.edu} \\
   \And
 Lazar ~Andjelic \\
  Department of Statistics\\
  Carnegie Mellon University\\
  Pittsburgh, PA 15213 \\
  \texttt{landjeli@andrew.cmu.edu} \\
  \And
 Amir ~Barati Farimani \\
  Assistant Professor of Mechanical Engineering\\
  Carnegie Mellon University\\
  Pittsburgh, PA 15213 \\
  \texttt{barati@cmu.edu} \\
}

%% file: Sections/abstract.tex
\begin{abstract}
Actor critic methods with sparse rewards in model-based deep reinforcement learning typically require a deterministic binary reward function that reflects only two possible outcomes: if, for each step, the goal has been achieved or not. Our hypothesis is that we can influence an agent to learn faster by applying an external environmental \textit{pressure} during training, which adversely impacts its ability to get higher rewards. As such, we deviate from the classical paradigm of sparse rewards and add a uniformly sampled reward value to the baseline reward to show that (1) sample efficiency of the training process can be correlated to the adversity experienced during training, (2) it is possible to achieve higher performance in less time and with less resources, (3) we can reduce the performance variability experienced seed over seed, (4) there is a maximum point after which more \textit{pressure} will not generate better results, and (5) that random positive incentives have an adverse effect when using a negative reward strategy, making an agent under those conditions learn poorly and more slowly. These results have been shown to be valid for Deep Deterministic Policy Gradients using Hindsight Experience Replay in a well known Mujoco environment, but we argue that they could be generalized to other methods and environments as well.
\end{abstract}

%% file: Sections/keywords.tex
\keywords{Deep Reinforcement Learning \and random reward \and POMDP \and DDPG \and HER}

%% file: Sections/introduction.tex
\section{Introduction}
\label{sec:introduction}

\begin{figure}[ht]
\centering
\includegraphics[width=1.0\linewidth]{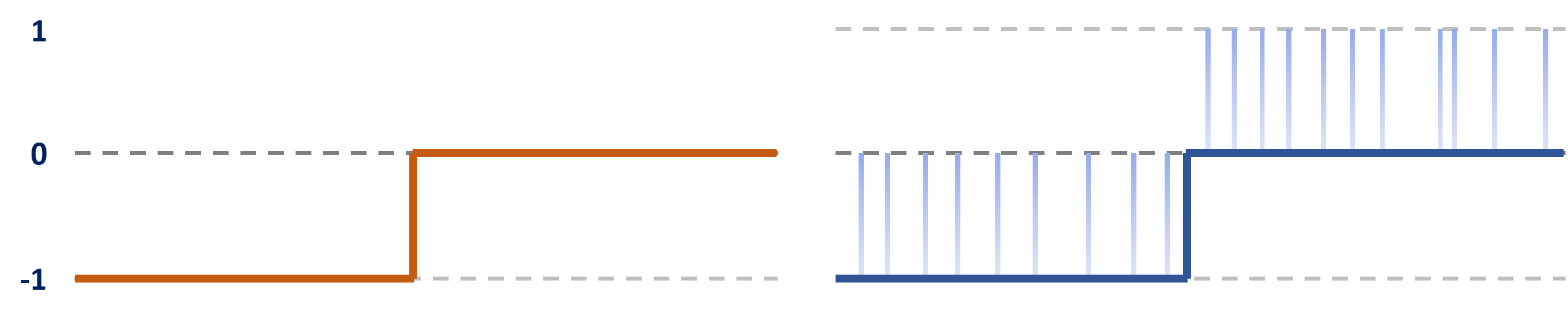}
\caption{\textbf{Left:} normal deterministic reward policy of $-1$ for every time the goal has not being achieved and 0 when it has. In this case, the cumulative expected reward is $-30$ assuming it takes half the time to complete the task. \textbf{Right:} the same agent receives additional uniformly sampled $+1$ rewards 50\% of the times. Assuming the agent again takes half the time to complete the task, the expected cumulative reward becomes $0$. This strategy in particular creates a learning problem. Throughout this paper many other strategies will be used that have the opposite effect.}
\label{fig:rr_step}
\end{figure}

The reward function in Deep Reinforcement Learning enables the agent to learn from the environment by providing feedback on the actions that it executes. Rewards can be dense, or they can be sparse. Rewards can also be originated by the environment (extrinsic), or be generated by the agent itself (intrinsic). When working with extrinsic sparse rewards in actor critic architectures, the decision of whether to use positive or negative rewards is addressed early on in the engineering process. There are works that seek to answer the question of which one to use. For instance, HER \cite{bib:HER} has used negative sparse rewards in the past leading to interesting results in robotics by forcing the robot to try to reach the goal as quickly as possible, then curiosity driven experiments \cite{bib:lanier2019curiositydriven} have used incremental intrinsic sparse rewards for achieving goals of increasing complexity. However, the magnitude of a single-task reward is not commonly addressed. After all, working with a value of -1 or -10, or any other negative number in this case, becomes an arbitrary reference point for the algorithm to understand that something undesirable is happening. So far, the premise of extrinsic sparse rewards has been to create a reward function that outputs one of two possible values or states in a deterministic fashion. One for a when a desirable outcome is reached, and the other at any other time (any other state that is not desirable automatically becomes undesirable). We propose the creation of a third state that is to be applied only during training that is meant to improve performance and sample efficiency. The third state, unlike the other two, is allowed to have a stochastic nature. When enabled, an additional constant reward, a \textit{bonus}, will be uniformly sampled throughout an episode and may occur at one or multiple steps either when the goal has not been achieved, when the goal has been achieved, or both (Fig \ref{fig:rr_step}). The significance of the third state and its effects on the learning process is the subject of the study of this paper.

\subsection{Partially Observable Markov Decision Processes}

This work is partially inspired by the subject of Partially Observable Markov Decision Processes (POMDPs) \cite{bib:POMDP} in which the observation made by the agent cannot fully explain the reason a reward is being awarded at all times. And, to a certain extent, the subject of Corrupt Reward Markov Decision Processes (CRMDPs) \cite{bib:CRMBP}, in which a reward function is corrupted by an external influence for which additional algorithms are created to detect and recover the original unaltered signal in order to achieve a better learning experience \cite{bib:spiky_CRMDPs}. Our work is closer to the theory of POMDPs than CRMDPs, however we are neither attempting to recover a Hidden Markov Model or recover the unaltered reward function. Instead, we seek to explore the agent behavior under conditions in which the reward function is not always consistent, or its magnitude is higher or lower during training. For this purpose, the uniform distribution plays a central role in ensuring that the signal cannot be learned and exploited by the agent. It also plays a central role in our ability to carry out the experiment and influence the agent into achieving a higher, mostly because we can increase the reward density more gradually and analyze the results. 

\subsection{Relationship to regularization in supervised learning}

While is not the main objective of this experiment, adding a stochastic component to the reward function may help create a more robust method to train model-free deep reinforcement learning agents because the randomly sampled values create a barrier to overfitting to a specific policy that maximizes the reward function. This line of research is not new. Previous research has shown the benefits of adding noise to the action space to be able to generalize better \cite{bib:dexterous_hand}, or using doing domain randomization, for instance in the field of computer vision, to enable robots to perform the bulk of their training in a simulated environment and yet perform well in real life conditions \cite{bib:Ren_2019}. We believe that we achieve a similar effect with the nature of our experiment.

\subsection{Terminology}

For the purpose of this paper, the terms \textit{robot} and \textit{agent} will be used interchangeably and mean the same thing. At times we refer to \textit{pressure} and \textit{adversity} as a way to indicate that negative rewards are being used in addition to the deterministic reward generated by the environment. These additional reward will, in general, also be called a \textit{bonus reward} or simply a \textit{bonus}, and its value can be positive or negative. The value of the bonus reward is not random, but the sampling of it is. Both the value of the bonus reward and the frequency of the random sampling are deterministic hyper-parameters that will be explained in later sections.

In simpler terms, the \textit{reward} s comprised of the baseline \textit{original reward} and a \textit{bonus}. In our experiments the \textit{original reward} follows the approach given in the HER paper \cite{bib:HER}.

%% file: Sections/methods.tex
\section{Methodology and Tools}
\label{sec:methods}

We are employing variations of the FetchSlide environment originally created and made available by OpenAI \cite{bib:OpenAIRoboEnv}, trained using Deep Deterministic Policy Gradients (DDPG) \cite{bib:DDPG} with the original version of Hindsight Experience Replay (HER) \cite{bib:HER} as made available within the OpenAI Baselines \cite{bib:baselines}. The FetchSlide environment is ideal for our research as it is a complex environment with a very specific set of actions that must to be taken in order for the agent to reach the goal and accumulate rewards. Therefore, modifying this environment in the way we describe below can be used to show the effects of random rewards in the learning process. The experiments were run in two different computers, each equipped with an Intel Core i7-8700K 3.70GHz $\times$ 12 cores CPU, 16 Gb of RAM, and a GeForce GTX 1080 Ti/PCIe/SSE2 GPU, running Ubuntu 16.06.6 LTS. For repeatibility purposes, both servers ran the same version of the software, including Python 3.7.3 with Tensorflow 1.13.1 and Mujoco 2.0. Each training sequence was executed on 2 cores for a total duration of 96,000 episodes. Testing sequences were 1,000 episodes long, and run on pre-trained robots. We verified that experiments running the same seed generate the same results on all versions of the environment.

\subsection{Description of the Environment}

The FetchSlide environment was specifically modified to generate a different set of rewards. The original reward function could be expressed as: 
\begin{equation} \label{eq:1}
    R: S \times A \times G \rightarrow \mathds{R}
\end{equation}
for which $R\left(s,ag,g\right) = 0$ when the ($x,y,z$) coordinates of the achieved goal ($ag$) are within the tolerance distance to the ($x,y,z$) coordinates of the goal ($g$), and -1 otherwise. We introduce:
\begin{equation} \label{eq:2}
    R': R \times P \times B \times N \rightarrow \mathds{R}
\end{equation}
for which $R$ is the set of original rewards, $P$ is the set of probability levels to obtain the nominal reward without any random bonuses. $B$ is the set of additional rewards, or bonuses, that the agent is subject to, in which B $\subseteq$ $\mathds{Z}$. And $N \rightarrow \{NG,G,B\}$ is the stage within each episode, for which the bonus $B$ is applied. We define $NG$ as the time span when the goal has not being achieved, $G$ as the time span while the goal has been achieved, and $B$ as the union of both, which is also equivalent to the entire duration of the episode. $P$, $B$, and $N$ are all deterministic hyperparameters for the experiment. Table \ref{tab:experiments} shows the list of experiments that were run. 

\begin{longtable}[c]{@{}lll@{}}
\caption{List of experiments}
\label{tab:experiments}\\
\toprule
Probability (P) & Bonus (B) & Stage (N) \\* \midrule
\endfirsthead
\multicolumn{3}{c}%
{{\bfseries Table \thetable\ continued from previous page}} \\
\toprule
Probability (P) & Bonus (B) & Stage (N) \\* \midrule
\endhead
\bottomrule
\endfoot
\endlastfoot
{[}0,10,30,50,70,90{]} & -15 & NG \\
{[}0,10,30,50,70,90{]} & -10 & NG \\
{[}0,10,30,50,70,90{]} & -5 & NG \\
{[}0,10,20,30,50,60,70,80,90{]} & -1 & B \\
{[}0,10,30,50,60,70,80,90{]} & -1 & G \\
{[}0,10,30,50,60,70,80,90{]} & -1 & NG \\
{[}{]} & 0 & HER Reference \\
{[}0,10,20,30,40,50,60,70,80,90{]} & 1 & B \\
{[}0,10,30,50,60,70,80,90{]} & 1 & G \\
{[}0,10,30,50,60,70,80,90{]} & 1 & NG \\
{[}0,10,30,50,70,90{]} & 10 & G \\* \bottomrule
\end{longtable}

The pertinent modifications to the environment are made to the function to compute the rewards. We first define $x\sim$ $N(0,1)$, sampled on each step, and generate the reward as the function as $r + b$ when $x \geq p$, and $r$ otherwise. Because of the way HER works, training typically occurs with a mix of about 20\% of the transitions having the environmental reward and the remaining 80\% of the transitions are modified to have a reward of $0$, as described by the HER algorithm \cite{bib:HER}. The mix ratio is a hyperparameter and can be modified, but for the purpose of our experiments it wasn't. Over time, the actual expected value of the reward for each step during training becomes $r + b(1 - p)$. 

Notice that the modified reward function calculation is only applied to \emph{extrinsic} rewards, those coming from the environment, and not those modified using the HER portion of the algorithm. These HER transitions are essential for baseline learning and, from the point of view of the experiment, act as \emph{intrinsic} rewards that serve the very specific purpose of learning from failed experiences, an outcome that we did not want to alter. 

Putting equations \ref{eq:1} and \ref{eq:2} together we get $R'\left(R\left(s,a,g\right),b,p,n\right)$. Because, as stated earlier, about 80\% of the experience replay buffer sampled transitions are used for HER and the remaining 20\% are used in accordance to the DDPG algorithm, we can calculate what the new average rewards for these states over time would be. When the goal has not been achieved (NG) and the standard reward is $-1$, the expected average value of the non-goal (NG) reward over time is $-1 + b(1 - p)$. When the goal has been achieved (G) and the standard reward is $0$, the expected average value over time for the goal reward is $0.2(r + b - pb) + 0.8r = 0.2b(1 - p)$. This shows that the randomness of the reward function has much more significance on the NG stage compared to the goal reward as a result of the intrinsic incentives generated by HER sampling. This difference in significance also affects how the critic network in DDPG approximates the Q function on the reward, as this approximation will be affected separately between transitions that led to the goal and transitions that did not, which changes how the transitions on either stage are valued, and can either greatly help or hurt the actor's policy.

\subsection{Testing}
 Testing uses the unmodified version of the FetchSlide environment that only generates deterministic rewards. For comparability purposes, all training experiments were performed using a common random seed, while testing was done using 5 random seeds not used during training.

%% file: Sections/results.tex
\section{Results}
\label{sec:results}

\begin{figure}[ht]
\centering
\includegraphics[width=1.0\linewidth]{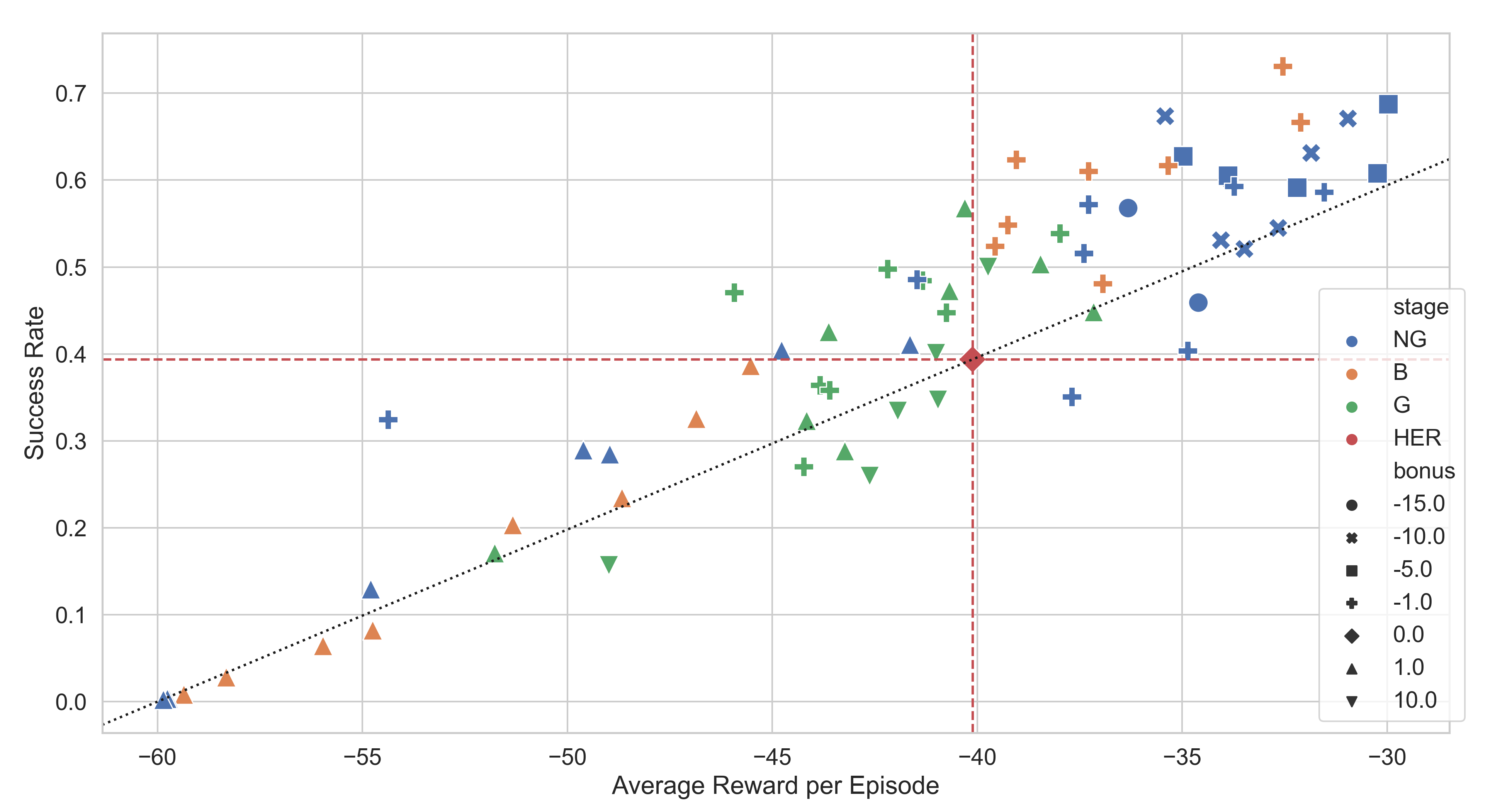}
\caption{Relationship between rewards and success rate during testing. Robots trained with more punitive rewards tend to have a higher success rate and higher rewards during testing (upper right quadrant). Conversely, robots trained with more lenient rewards tend to perform worse during testing.}
\label{fig:rr_rew_vs_sr_testing}
\end{figure}

\begin{figure}[ht]
\centering
\includegraphics[width=1.0\linewidth]{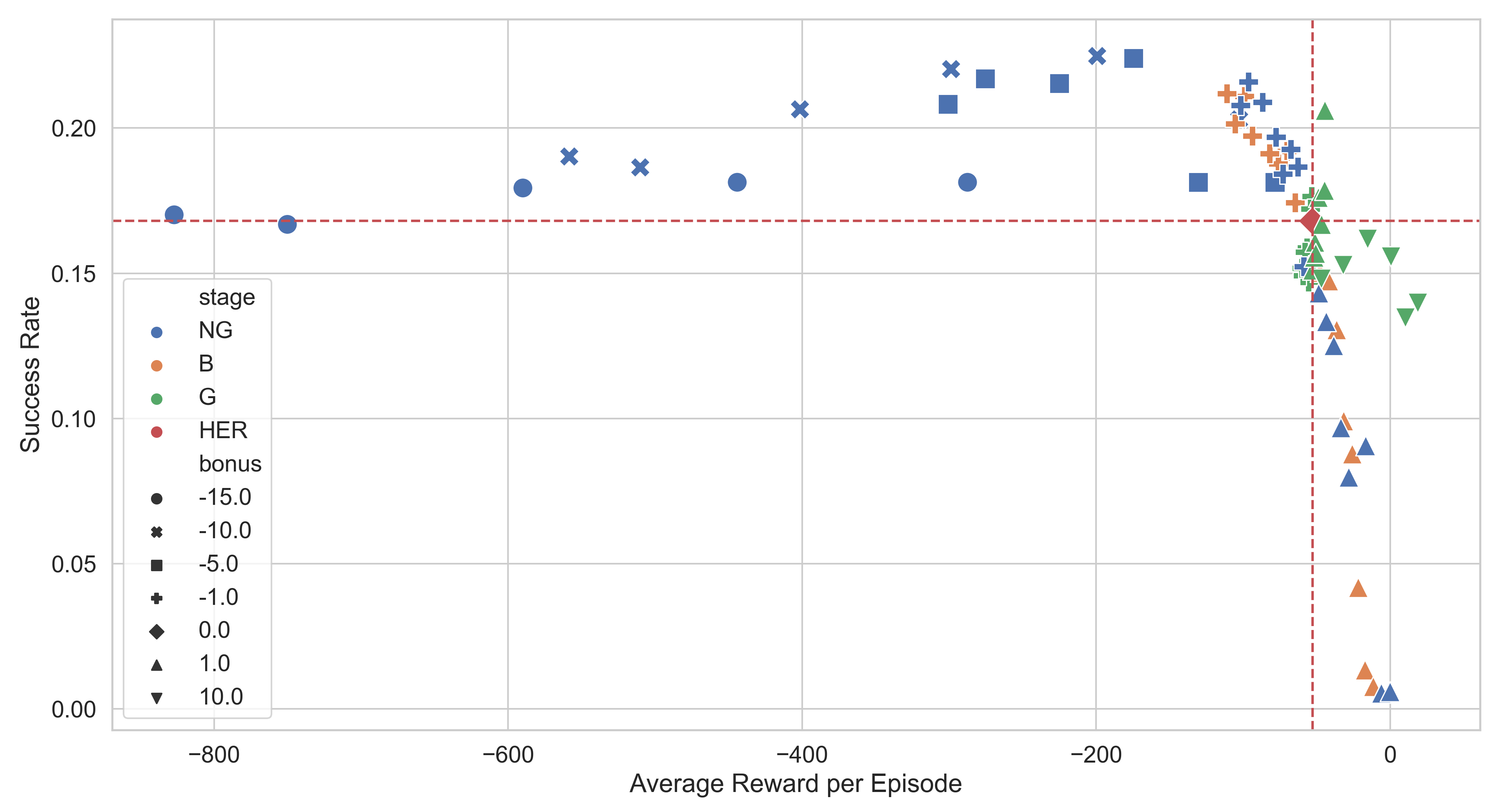}
\caption{Relationship between rewards and success rate during training. Robots trained under an environment that feeds more negative random rewards (upper left quadrant) tend to outperform those that trained under more positive random rewards (lower right quadrant). The red markers represent the baseline result from a comparable training process with no random rewards. }
\label{fig:rr_rew_vs_sr_training}
\end{figure}

The results show that it is possible to modify the robot's performance in a predictable way by applying our methodology during training. The results show that more negative sparse rewards induce a better performance in both maximizing the accumulated reward (the objective function), and maximizing the success rate. While typically these two, rewards and success rate, are correlated over time, for this experiment we can see how the correlation changes when modifying the configuration of the reward (fig. \ref{fig:rr_rew_vs_sr_training}). Interestingly enough, there is a threshold point after which achieving better results becomes harder, and augmenting the absolute value of the negative reward will no longer improve performance. The data also shows that there are several paths to achieve a higher performance. For instance, [100:-1:B] seems to generate one of the best results, but so does [50:-5:NG] or [30:-10:NG]. The reference point, showed in red (fig. \ref{fig:rr_rew_vs_sr_training}) corresponds to the nominal HER results. A dotted line runs through it from the origin to show that the success rate was seen to improve in general more so than the rewards. Positive bonuses, on the other hand, perform worse. In a sense, this performance is expected. It's important to remember that we are using a negative reward strategy [-1,0]. As such, positive bonuses should be understood as rewards that cancel out the effects of the underlying reward strategy to the point in which is not clear whether an action is good or bad. 

\begin{figure}[ht]
\centering
\includegraphics[width=0.9\linewidth]{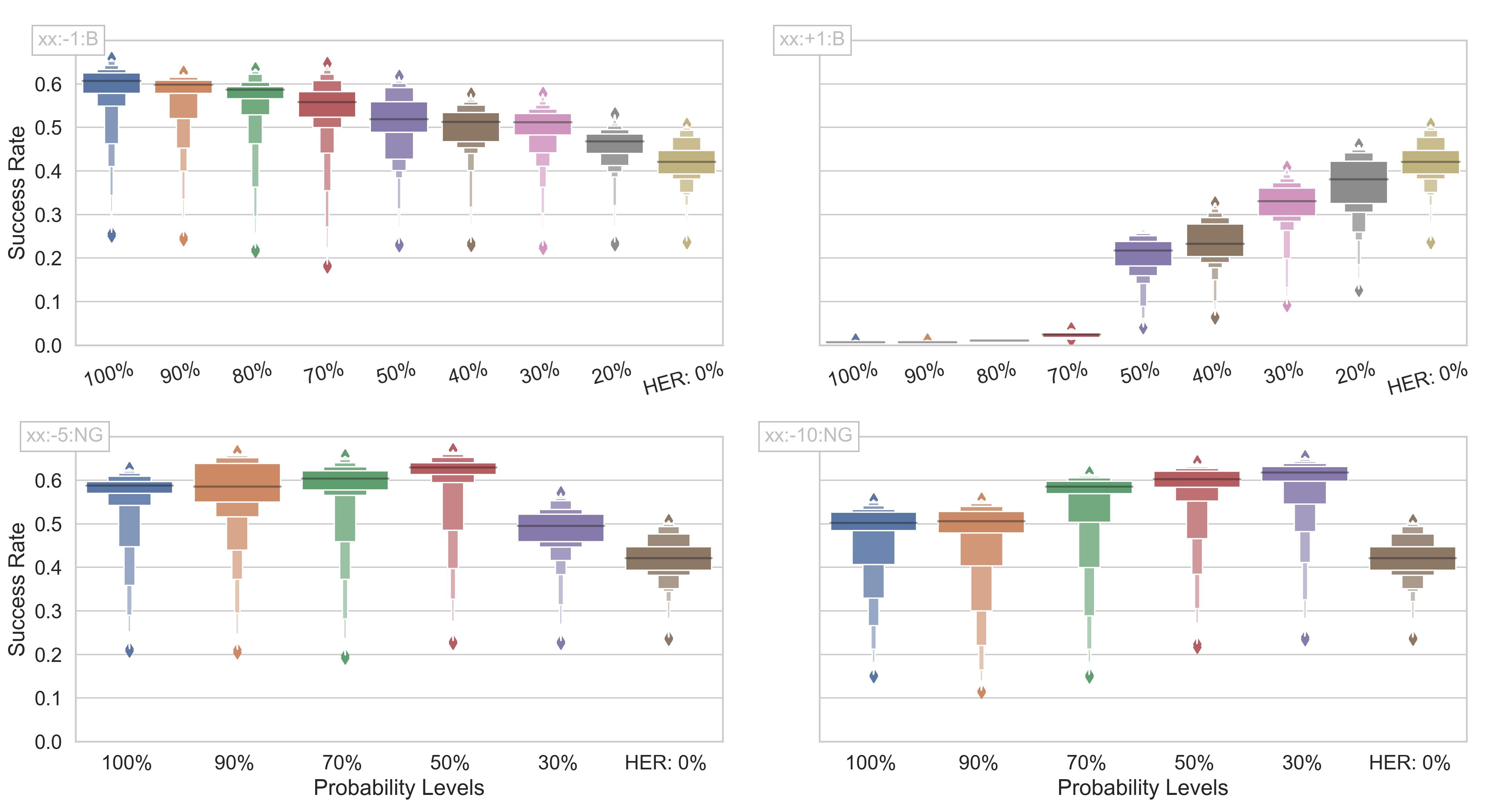}
\caption{\textbf{Upper left:} the [xx:-1:B] series shows progressive improvement as the probability of having additional bonus rewards of -1 at each time step increases, while the \textbf{Upper right:} [xx:+1:B] series shows the analogous effects for a positive reward. After the probability jumps above 50\%, the robot arguably stops learning. \textbf{Lower left:} the [xx:-5:NG] series shows what happens when the bonus is further reduced: the improvement stops and starts to slowly descend. \textbf{Lower right:} the [xx:-10:NG] series shows a similar pattern in which more negative rewards start to degrade performance, while at the same time showing that a few more negative rewards can have a similar effect to a lot of less negative rewards.}
\label{fig:rr_prob_vs_sr_1B_testing}
\end{figure}

The case for which the robot performs the worst is when the bonus is +1 (fig. \ref{fig:rr_prob_vs_sr_1B_testing} upper right), which makes sense as it would fully overlap with the original [-1,0] reward corresponding to the [NG:G] stages. This slowly transforms the rewards into a [0:+1] model without modifying the HER reward, which is fixed at 0. We argue that the learning process wouldn't be as poor if we changed the HER reward to +1. That said, one of the best performances comes from using the opposite bonus of -1 (fig. \ref{fig:rr_prob_vs_sr_1B_testing} upper left), getting incrementally better as the probability of sampling an additional bonus reward on each time step goes higher. When the probability reaches 100\% the reward set becomes [-2, -1] for the [NG, G] stages, and 0 for HER. When using an even lower bonus value of -5 (fig. \ref{fig:rr_prob_vs_sr_1B_testing} lower left), the success rate starts to increase as it did with -1, but at around a probability level of 50\% it reaches a maximum (actually, the best result among all experiments) and then the performance starts to slowly deteriorate. For the purpose of this paper, we didn't seek to find the exact bonus/probability combination at which we can obtain the best possible performance, but instead we were interested in understanding the general dynamics of the process. As such, we also performed an experiment with even lower bonus rewards, such as -10 (fig. \ref{fig:rr_prob_vs_sr_1B_testing} lower right). At this bonus level, the lowest probability we tried (30\%) yielded the best result for the series and the second best for the entire project. This result gives credence to the statement we made before about there being several strategies to achieve peak performance, and one of them may very well be having a low probability at all times of getting a relatively lower reward for no good reason (random sampling). This result also suggests something that at this point should be obvious. That having multiple levels of sparse rewards could be beneficial under certain conditions, even when they are not linked with the specific environmental outcomes (bonus rewards are randomly sampled following a given uniform distribution). This detachment from the environment is important, since linking sparse reward levels to environmental outcomes would be akin to engineering rewards, which requires significant time and domain expertise.

During testing, we run the trained models using the regular HER algorithm, absent any bonus rewards. It is only during training that the bonus rewards are applied (which is why on previous figures the minimum reward is -60 for each 60 time step episode). However, the training results do provide an interesting insight into what is going on. For instance, the top performers in terms of the success rate during testing, are also the top performers during training, even if they collect much lower rewards because of the conditions we artificially subject the agent to (fig. \ref{fig:rr_rew_vs_sr_training}). Also, the success rate is much lower during training because it averages the entire learning process from zero. That being said, the data does suggest that there is a maximum performance that can be achieved for this specific agent, environment, hyperparameters; and other conditions, physical or otherwise, associated with our test bed. That peak performance can be achieved when the average reward per episode during training is around -200. This value is irrelevant in absolute terms, but it is indicative of how much performance we can expect to extract.

%% file: Sections/conclusion.tex
\section{Conclusion}
\label{sec:conclusion}

From these experiments we conclude that it is possible to improve the agent's performance and sample efficiency by structuring a partially deterministic sparse reward function such as the ones we applied in this paper. By doing so, we would more closely approximate the performance boundaries the agent in that environment is capable of reaching, for any given seed. 

\subsection{Future work}
We didn't analyze the effects of having multiple bonus levels at multiple probabilities at the same time. Such a result would more closely resemble real life, in which the environment is not always in agreement about what reward should be given at any given time. As valuable as that would be, such a test would require a much larger test bed and computing resources to draw conclusions about the right combination of factors. We would, however, expect to see a change in performance similar to what we have shown in our experiments. We also anticipate that performance deterioration would be caused by receiving random bonus rewards with a magnitude that is too high, or bonus rewards that make the distinction between when the goal has been achieved and when it hasn't less clear.
Replicating these results on other model-free deep reinforcement learning algorithms is also something we can see happening in the future.

%% file: Sections/supplemental.tex
\newpage
\renewcommand{\thefigure}{SI-\arabic{figure}}
\setcounter{figure}{0}
\renewcommand{\thetable}{SI-\arabic{table}}
\setcounter{table}{0}
\section*{Supplemental Information}
\label{sec:supplemental}

\begin{figure}[ht]
\centering
\includegraphics[width=1.0\linewidth]{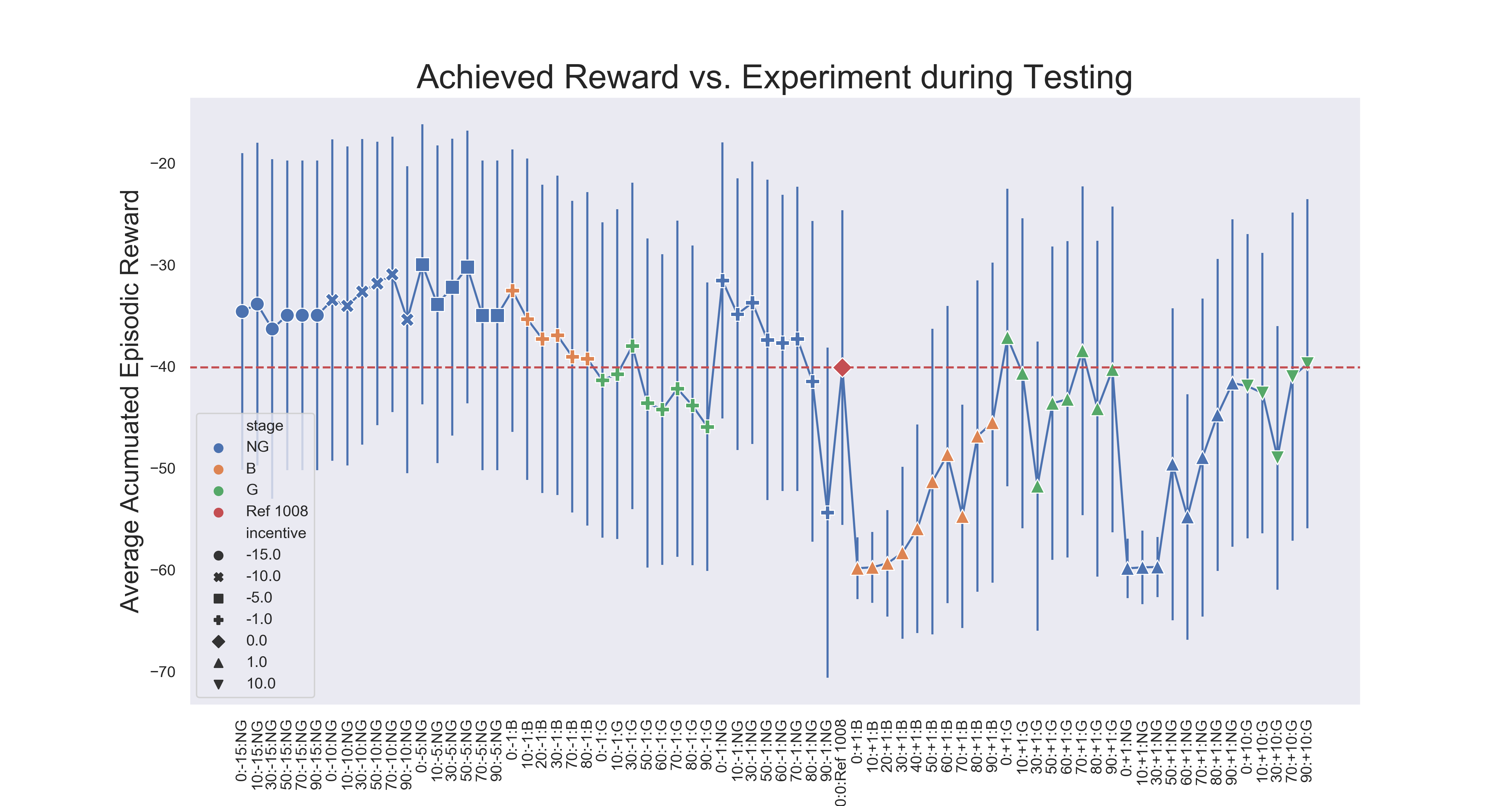}
\caption{Average reward per experiment conducted clearly shows the performance difference between the different configurations. The naming convention is P:B:N, in which the P represents the percentage of rewards that maintain the original deterministic reward, B represents the numerical amount that is added to the deterministic reward, and N represents whether for each episode the additional reward B applies to when the robot has not achieved the goal (NG), has achieved it (G), or both (B). Bars reflect the standard deviation.}
\label{fig:rr_rew_vs_exp_testing}
\end{figure}

Identifying the best combination of parameters to achieve the best possible results for a robot is important. In the case of our experiments, we show how the combination of stage, probability and bonus affect the average outcome of the experiments (fig \ref{fig:rr_rew_vs_exp_testing}). The maximum performance is achieved when fully offsetting the rewards by -1 (0:-1:NG), applying various combinations of random rewards of value -5, or very sparsely applying random rewards of -10 (70:-10:NG). At the same time, positive rewards should almost entirely be avoided, especially fully offsetting the reward by +1 (0:+1:B or 0:+1:NG), at which point the robot will encounter serious problems learning a working policy. The full table of results is also presented below (Table \ref{tab:results})

\begin{longtable}[c]{@{}llllllll@{}}
\caption{Experimental Results}
\label{tab:results}\\
\toprule
Id & Probability & Bonus & Stage & Test reward & Test Success Rate & Train reward & Train Success Rate \\* \midrule
\endfirsthead
\multicolumn{8}{c}%
{{\bfseries Table \thetable\ continued from previous page}} \\
\toprule
Id & Probability & Bonus & Stage & Test reward & Test Success Rate & Train reward & Train Success Rate \\* \midrule
\endhead
\bottomrule
\endfoot
\endlastfoot
1 & 0 & -15 & NG & -34.607 & 0.459 & -827.155 & 0.170 \\
2 & 10 & -15 & NG & -33.875 & 0.602 & -750.148 & 0.167 \\
3 & 30 & -15 & NG & -36.319 & 0.568 & -589.953 & 0.179 \\
4 & 50 & -15 & NG & -34.976 & 0.628 & -444.147 & 0.181 \\
5 & 70 & -15 & NG & -34.976 & 0.628 & -287.494 & 0.181 \\
6 & 90 & -15 & NG & -34.976 & 0.628 & -130.718 & 0.181 \\
7 & 0 & -10 & NG & -33.482 & 0.521 & -558.401 & 0.190 \\
8 & 10 & -10 & NG & -34.055 & 0.531 & -510.140 & 0.186 \\
9 & 30 & -10 & NG & -32.659 & 0.545 & -401.488 & 0.206 \\
10 & 50 & -10 & NG & -31.855 & 0.631 & -298.677 & 0.220 \\
11 & 70 & -10 & NG & -30.957 & 0.671 & -199.305 & \textbf{0.225} \\
12 & 90 & -10 & NG & -35.416 & 0.674 & -102.720 & 0.203 \\
13 & 0 & -5 & NG & \textbf{-29.977} & 0.688 & -300.759 & 0.208 \\
14 & 10 & -5 & NG & -33.890 & 0.605 & -275.362 & 0.217 \\
15 & 30 & -5 & NG & -32.201 & 0.591 & -225.137 & 0.215 \\
16 & 50 & -5 & NG & -30.237 & 0.608 & -174.574 & 0.224 \\
17 & 70 & -5 & NG & -34.976 & 0.628 & -130.667 & 0.181 \\
18 & 90 & -5 & NG & -34.976 & 0.628 & -78.409 & 0.181 \\
19 & 0 & -1 & B & -32.544 & \textbf{0.731} & -111.100 & 0.212 \\
20 & 10 & -1 & B & -35.351 & 0.617 & -105.442 & 0.201 \\
21 & 20 & -1 & B & -37.293 & 0.610 & -99.096 & 0.211 \\
22 & 30 & -1 & B & -36.943 & 0.481 & -93.636 & 0.197 \\
23 & 50 & -1 & B & -32.111 & 0.667 & -81.983 & 0.191 \\
24 & 60 & -1 & B & -39.571 & 0.524 & -76.142 & 0.188 \\
25 & 70 & -1 & B & -39.047 & 0.624 & -70.141 & 0.192 \\
26 & 80 & -1 & B & -39.255 & 0.549 & -64.564 & 0.174 \\
27 & 0 & -1 & G & -41.340 & 0.484 & -60.000 & 0.152 \\
28 & 10 & -1 & G & -40.761 & 0.447 & -59.364 & 0.149 \\
29 & 30 & -1 & G & -37.990 & 0.539 & -57.963 & 0.157 \\
30 & 50 & -1 & G & -43.604 & 0.358 & -56.692 & 0.159 \\
31 & 60 & -1 & G & -44.236 & 0.270 & -56.151 & 0.154 \\
32 & 70 & -1 & G & -42.190 & 0.498 & -55.584 & 0.147 \\
33 & 80 & -1 & G & -43.838 & 0.364 & -54.139 & 0.172 \\
34 & 90 & -1 & G & -45.931 & 0.471 & -53.556 & 0.176 \\
35 & 0 & -1 & NG & -31.542 & 0.586 & -101.709 & 0.208 \\
36 & 10 & -1 & NG & -34.867 & 0.404 & -96.383 & 0.216 \\
37 & 30 & -1 & NG & -33.735 & 0.593 & -86.827 & 0.209 \\
38 & 50 & -1 & NG & -37.399 & 0.516 & -77.609 & 0.197 \\
39 & 60 & -1 & NG & -37.695 & 0.351 & -72.964 & 0.184 \\
40 & 70 & -1 & NG & -37.286 & 0.572 & -67.660 & 0.193 \\
41 & 80 & -1 & NG & -41.469 & 0.486 & -62.779 & 0.187 \\
42 & 90 & -1 & NG & -54.375 & 0.324 & -58.921 & 0.152 \\
43 & 100 & 0 & REF & -40.115 & 0.394 & -52.914 & 0.168 \\
44 & 0 & 1 & B & -59.843 & 0.003 & 0.362 & 0.006 \\
45 & 10 & 1 & B & -59.751 & 0.003 & -5.613 & 0.006 \\
46 & 20 & 1 & B & -59.354 & 0.008 & -11.437 & 0.008 \\
47 & 30 & 1 & B & -58.320 & 0.028 & -17.051 & 0.013 \\
48 & 40 & 1 & B & -55.959 & 0.064 & -21.658 & 0.042 \\
49 & 50 & 1 & B & -51.329 & 0.203 & -25.752 & 0.088 \\
50 & 60 & 1 & B & -48.666 & 0.234 & -31.555 & 0.099 \\
51 & 70 & 1 & B & -54.747 & 0.082 & -36.357 & 0.131 \\
52 & 80 & 1 & B & -46.853 & 0.326 & -41.731 & 0.147 \\
53 & 90 & 1 & B & -45.529 & 0.386 & -47.654 & 0.150 \\
54 & 0 & 1 & G & -37.160 & 0.448 & -44.568 & 0.179 \\
55 & 10 & 1 & G & -40.678 & 0.473 & -46.625 & 0.167 \\
56 & 30 & 1 & G & -51.770 & 0.171 & -44.343 & 0.206 \\
57 & 50 & 1 & G & -43.623 & 0.425 & -48.756 & 0.176 \\
58 & 60 & 1 & G & -43.229 & 0.288 & -50.608 & 0.157 \\
59 & 70 & 1 & G & -38.456 & 0.503 & -51.124 & 0.161 \\
60 & 80 & 1 & G & -44.161 & 0.323 & -51.679 & 0.155 \\
61 & 90 & 1 & G & -40.302 & 0.568 & -52.806 & 0.151 \\
62 & 0 & 1 & NG & -59.853 & 0.002 & 0.000 & 0.006 \\
63 & 10 & 1 & NG & -59.756 & 0.003 & -5.963 & 0.005 \\
64 & 30 & 1 & NG & -59.718 & 0.003 & -16.593 & 0.091 \\
65 & 50 & 1 & NG & -49.609 & 0.289 & -28.109 & 0.080 \\
66 & 60 & 1 & NG & -54.794 & 0.129 & -33.473 & 0.097 \\
67 & 70 & 1 & NG & -48.963 & 0.285 & -38.338 & 0.125 \\
68 & 80 & 1 & NG & -44.772 & 0.404 & -43.406 & 0.133 \\
69 & 90 & 1 & NG & -41.641 & 0.411 & -48.454 & 0.143 \\
70 & 0 & 10 & G & -41.939 & 0.334 & 18.850 & 0.140 \\
71 & 10 & 10 & G & -42.624 & 0.260 & 10.402 & 0.135 \\
72 & 30 & 10 & G & -48.986 & 0.157 & 0.585 & 0.156 \\
73 & 50 & 10 & G & -40.958 & 0.347 & -15.278 & 0.162 \\
74 & 70 & 10 & G & -41.006 & 0.401 & -31.890 & 0.153 \\
75 & 90 & 10 & G & -39.736 & 0.500 & -46.748 & 0.148 \\* \bottomrule
\end{longtable}